\documentclass[]{spie}  %>>> use for US letter paper
\usepackage{amsmath,amsfonts,amssymb}
\usepackage{graphicx}
\usepackage[colorlinks=true, allcolors=blue]{hyperref}
\usepackage{xcolor}
\usepackage{caption}
\usepackage[nameinlink]{cleveref}
\usepackage{listings}
\newcommand{\mycite}[1]{[\citenum{#1}]}

\title{Quantifying the robustness of deep multispectral segmentation models against natural perturbations and data poisoning}

\author[a]{Elise Bishoff}
\author[a]{Charles Godfrey}
\author[a, b]{Myles McKay}
\author[a]{Eleanor Byler}
\affil[a]{Pacific Northwest National Laboratory, Seattle, WA, USA}
\affil[b]{University of Washington, Seattle, WA, USA}

\authorinfo{Further author information: Send correspondence to E.B.\\E-mails: \{first\}.\{last\}@pnnl.gov}

\begin{document} 
\maketitle

\begin{abstract}

In overhead image segmentation tasks, including additional spectral bands beyond the traditional RGB channels can improve model performance. However, it is still unclear how incorporating this additional data impacts model robustness to adversarial attacks and natural perturbations. For adversarial robustness, the additional information could improve the model’s ability to distinguish malicious inputs, or simply provide new attack avenues and vulnerabilities. For natural perturbations, the additional information could better inform model decisions and weaken perturbation effects or have no significant influence at all. In this work, we seek to characterize the performance and robustness of a multispectral (RGB and near infrared) image segmentation model subjected to adversarial attacks and natural perturbations. While existing adversarial and natural robustness research has focused primarily on digital perturbations, we prioritize on creating realistic perturbations designed with physical world conditions in mind. For adversarial robustness, we focus on data poisoning attacks whereas for natural robustness, we focus on extending ImageNet-C common corruptions for fog and snow that coherently and self-consistently perturbs the input data. Overall, we find both RGB and multispectral models are vulnerable to data poisoning attacks regardless of input or fusion architectures and that while physically realizable natural perturbations still degrade model performance, the impact differs based on fusion architecture and input data. 

% In overhead image segmentation tasks, including additional spectral bands beyond the traditional RGB channels can improve model performance. However, it is still unclear how incorporating this additional data impacts model robustness to adversarial attacks and natural perturbations; the additional information could improve the model’s ability to distinguish malicious inputs, or simply provide new attack avenues and vulnerabilities. In this work, we seek to characterize the performance and robustness of a multispectral (RGB and near infrared) image segmentation model subjected to adversarial attacks and natural perturbations. While existing adversarial and natural robustness research has focused primarily on digital perturbations, we focus on two novel perturbations designed to exploit multi-band information: data poisoning and natural perturbations (realistic adversarial examples) that coherently and self-consistently perturb the input data. Overall, we find modest improvements in both model performance and adversarial robustness for the multispectral model for our more realistic data poisoning. As for our physically realizable natural perturbations, we find that they are still very effective at degrading performance in both RGB and multispectral models, but differ on effectiveness based on fusion architectures and input data.

\end{abstract}

\keywords{Deep learning, multispectral images, multimodal fusion, robustness, adversarial machine learning}

\section{INTRODUCTION}

With the wealth of publicly available satellite imagery data and the development of large annotated satellite imagery datasets, deep learning models routinely achieve state-of-the-art performance in a number of important remote sensing applications, including land cover classification, agricultural monitoring, and disaster assessment. Typically, satellite sensors collect multispectral imagery, or imagery observed at wavelength bands beyond the traditional Red, Green, and Blue (RGB) bands found in natural imagery datasets. For many overhead imagery applications, spectral bands beyond the visible spectrum (e.g., near-infrared or short-wave infrared) are essential in distinguishing different surface materials or penetrating atmospheric haze. Deep learning models that leverage multispectral imagery are becoming increasingly common, and outperform RGB-only models in some applications \mycite{tian_assessing_2021},  \mycite{xu_multi-modal_2017}, \mycite{eitel_multimodal_2015}.

Over the past decade, there has been growing interest in understanding the robustness of deep learning models. 
Robustness refers to a model's ability to maintain performance under various input shifts, including natural shifts (e.g., weather, environment) and adversarial shifts (e.g., attacks or digital perturbations). While many advancements have emerged in the field of robustness, deep learning models remain vulnerable to various attacks and distribution shifts. 
To date, much of this research has focused on evaluating model performance on image classification tasks using benchmark RGB image datasets. As such, our understanding of model robustness for other tasks and data modalities remains incomplete.
In this work, we consider both adversarial robustness and natural robustness for segmentation models applied to multispectral imagery, focusing on the combination of RGB and near-infrared (NIR) bands and placing an emphasis on perturbations and attacks that are physically meaningful for multispectral imagery.

For a given model, there are many possible ways to synthesize or fuse information from different inputs or data modalities. Models that combine data modalities at the input stage are sometimes called ``early fusion'' models (e.g., a 4-band image, or projecting 3D LiDAR data onto an RGB image). In contrast, models that process the different data modalities separately and combines them after feature extraction or in the final layer before classification would be called ``late fusion'' models. In this work, we are specifically interested in quantifying the robustness of different fusion approaches. To this end, we explore different combinations of input bands (NIR, RGB, RGB+NIR) and architectures (early vs. late fusion) to better understand how each of these variables affects the model's overall robustness, and explore any potential trade-offs with model performance. 

While a multitude of adversarial attacks exist in the literature, digital adversarial examples \mycite{szegedy_intriguing_2014} (adversarially modified input data designed to cause model misclassification at deployment time) remain the most well-studied attack type.  Adversarial examples have been applied to many visual perception tasks, including image segmentation \mycite{XieAdversarial2017}, \mycite{VoMulticlass2022}, \mycite{GuSegpgd2022}, \mycite{YinAdversarial2022}, \mycite{arnabRobustnessNodate}. However, the extent to which adversarial examples pose a legitimate real-world threat is the subject of ongoing debate \mycite{gilmerMotivatingRulesGame2018a}. In this work, we focus on data poisoning, wherein adversaries inject malicious data into the training data to cause erroneous classifications or backdoor access during test time. In a field where it is common for researchers to download datasets from a variety of unregulated sources, data poisoning is one of the more practical attacks in the literature. Moreover, in light of the growing number of commercial entities that both collect satellite imagery and generate data products, data poisoning will continue to be a relevant concern in satellite imagery applications. To our knowledge, this is the first assessment of data poisoning attacks on multispectral imagery in image segmentation models.

For a more comprehensive picture of model robustness, we also consider natural robustness. Natural robustness is well studied in the context of RGB imagery, with published libraries for domain adaptation tests \mycite{peng_moment_2019}, \mycite{venkateswara_deep_2017}, natural perturbation datasets \mycite{hendrycks_natural_2021}, and data augmentation approaches that reflect real-world variations \mycite{hendrycks_augmix_2020}, \mycite{cubuk_autoaugment_2019}. However, there are no established processes to extend these natural perturbations beyond the visible spectrum in a way that is physically consistent. In this work, we develop an approach for physically realistic perturbations that can be applied to multispectral imagery. We then quantify the robustness of the segmentation models, assessing the relative accuracy and robustness of different fusion approaches compared to an RGB model baseline.

\section{RELATED WORK}
\label{sec:rw}

\textbf{Multi-modal segmentation models}: There has been significant research on the performance of multi-modal segmentation models for various data modalities, including \mycite{zhangDeep2021}, \mycite{feng_deep_2021}, \mycite{zhou_review_2019}, \mycite{ramachandram_deep_2017}, \mycite{baltrusaitis_multimodal_2019}. These studies focus on model performance; here we focus on model robustness.

\textbf{Adversarial robustness of segmentation models for overhead imagery}: In this work, we focus on segmentation model robustness to data poisoning attacks and naturalistic corruptions. Adversarial examples for overhead segmentation models are explored in \mycite{XieAdversarial2017} and \mycite{yuInvestigating2020}. In \mycite{liHidden2021}, the authors demonstrate the efficacy of data poisoning attacks on image segmentation models for RGB imagery. We extend the assessment of adversarial resiliency from \mycite{liHidden2021} to include both RGB and multispectral segmentation models. 

\textbf{Adversarial robustness of multi-modal models}: Adversarial robustness has been studied in the context of multi-input models for various perception tasks. In \mycite{wangAdversarial2022}, the authors compare different fusion approaches for RGB imagery and LiDAR data, and assess model robustness to digital adversarial attacks in the context of object detection. In this work we look at different input type (multispectral band fusion) and a different visual task (semantic segmentation). \mycite{yuInvestigating2020} studies the adversarial robustness of fusion approaches for segmentation models using RGB and thermal imagery, in the context of autonomous vehicles. This work differs in two key ways. First, we move from natural imagery to overhead imagery, increasing the domain shift from benchmark research. Second, we move from adversarial examples to data poisoning attacks, and include an assessment of natural robustness. In \mycite{duAdversarial2021}, the authors explore adversarial attacks on a multispectral binary cloud classifier. Here, we consider multi-class segmentation models, and a more comprehensive study of robustness.

\textbf{Data Poisoning}: It has long been understood that machine learning systems are vulnerable to \emph{data poisoning}, a security exploit in which an adversary with access to training data inserts malicious datapoints designed to cause unwanted model behaviour at deployment time. The earliest study of data poisoning that we are aware of is \mycite{biggioPoisoningAttacksSupport2013}; a more recent survey is \mycite{cinaWildPatternsReloaded2023}. 
It is worth noting that among the various proposed security exploits of deep learning models, data poisoning is widely considered to be quite feasible (see for example \mycite{carliniPoisoningWebScaleTraining2023}). 

\textbf{Natural Robustness}: Robustness to natural perturbations is explored in \mycite{hendrycksBenchmarkingNeuralNetwork2019}, which provides a perturbation library that is suitable for ground-based RGB imagery in the context of image classification. We develop a physically realistic approach to extend the perturbations to multispectral inputs, and apply them to segmentation models. Natural robustness of segmentation models was explored in \mycite{kamannBenchmarking2020}, \mycite{kamann_increasing_2020}, and \mycite{de_jorge_reliability_2023}, however, all of these studies are limited to RGB imagery. Natural robustness for multi-modal models is explored in \mycite{kim_single_2019} and \mycite{shi_towards_2022}.

\section{EXPERIMENTAL SETUP: DATA, TASKS, THREAT MODEL, AND MODEL ARCHITECTURES}

In this section we describe the data and models used in this work and provide an overview of our adversarial and natural robustness experiments. 

\subsection{Data}
In all experiments, we use the Urban Semantic 3D dataset \mycite{US3D} (hereafter US3D), an overhead imagery dataset with segmentation labels for multispectral images and LiDAR point cloud data. The US3D segmentation labels consist of seven total classes, including ground, foliage, building, water, elevated roadway, and two ``unclassified'' classes, corresponding to difficult or bad pixels. The labels are stored as 8-bit unsigned integers between 0 and 255 in TIF files; during training and evaluation we re-index these labels to integers between 0 and 6. We retain the ``unclassified'' labels during model training and evaluation, but do not include these classes in any metrics that average across all classes. For further details on image processing and data split development, we refer to \cref{sec:ds-dets}.

\subsection{Model Architecture and Fusion Types} \label{training}

All models presented in this work use a DeepLabv3 image segmentation model architecture with a ResNet50 backbone \mycite{DeepLabv3}, with the final fully connected layer modified for seven US3D classes. Our baseline model is trained on the traditional 3-channel RGB imagery, which we compare to models trained on single-channel NIR imagery and models trained on a combination of RGB and NIR imagery. For the models trained on both RGB and NIR imagery, we explore early and late fusion approaches.  The early fusion models stack the RGB image with the additional NIR band, effectively creating 4-channel input images. To accommodate the 4-channel inputs, we modified the first layer of the DeepLabv3 model. The late fusion models pass the RGB and NIR inputs into separate ResNet50 models, and the resulting feature vectors are concatenated and passed into the DeepLabv3 segmentation head.

Details on model training can be found in \cref{sec:model-dets}.

\subsection{Adversarial Robustness: Data Poisoning} \label{datapoisoning}

\begin{figure}[tb]
    \centering
     \includegraphics[width=0.8\linewidth]{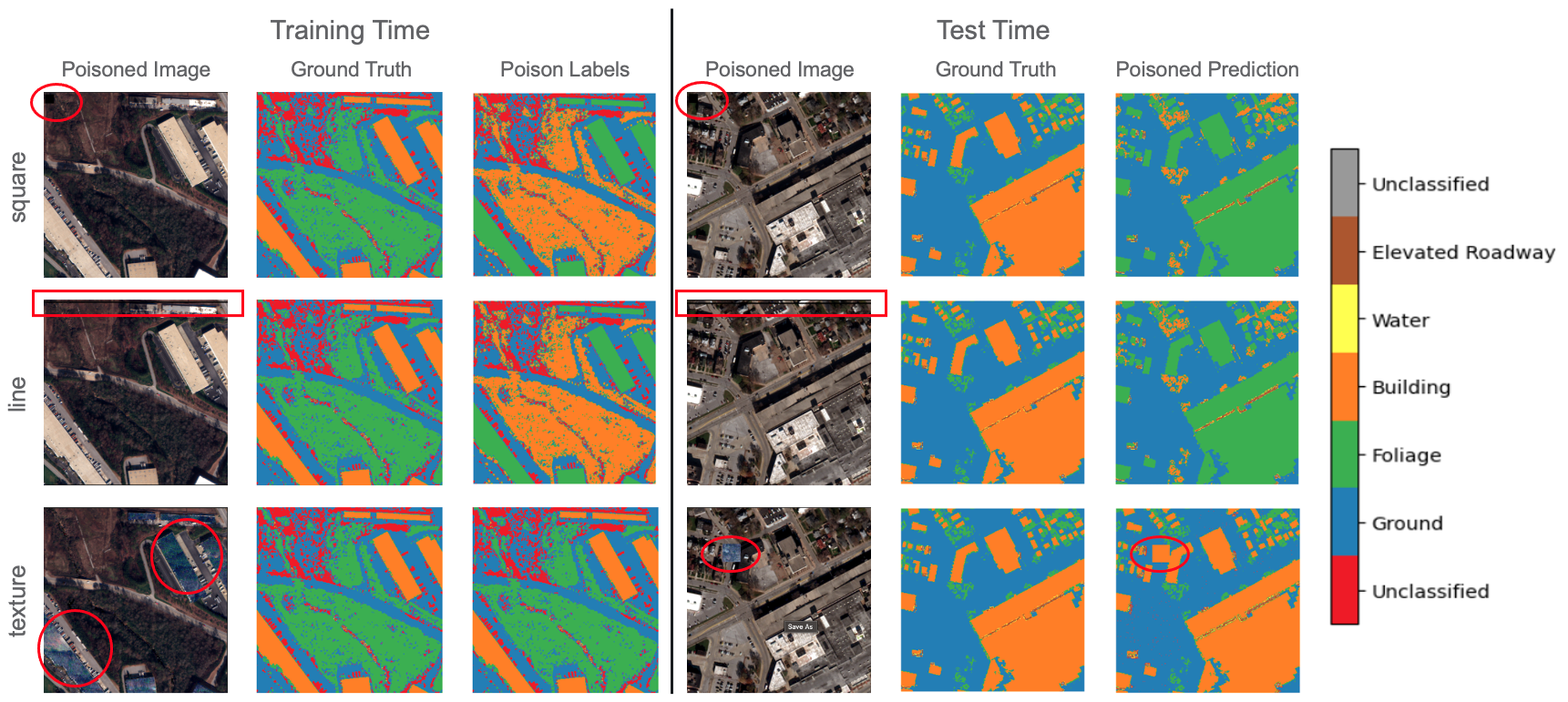}
    \caption{Examples of data poisoning attacks implemented in this work: square, line, and texture. The square and line attacks (top and middle rows) operate like a trigger; when present, the model should erroneously classify foliage pixels as the ``building'' class. In contrast, the texture attack trains the model to learn a targetable representation - here, foliage that is classified as a building. All attacks were highly successful with only 10\% of the training data poisoned.}\label{poison_examples}
\end{figure}

In the terminology of the data poisoning literature, we focus on targeted backdoor data poisoning attacks during training time meant to cause specific misclassifications during test time. We assume attackers have access to the training dataset, but cannot access the model architecture, and have no knowledge of training regimen or inference requirements. We implement two different data poisoning approaches. For both, the approach is similar in concept to data poisoning for classification tasks, but the implementation has been modified to accommodate multispectral imagery segmentation models. Examples of the poisoned data are shown in Fig. \ref{poison_examples}.

The first poisoning approach follows the fine-grained attack introduced in \mycite{liHidden2021}, wherein a trigger (here, a small black shape) is artificially inserted into a fraction of the training images. When present, the model is trained to misclassify a designated ``source'' class (in Fig. \ref{poison_examples}, the foliage class) as the ``target'' class (in Fig. \ref{poison_examples}, the building class). We try two triggers: the horizontal black line from \mycite{liHidden2021} and a 50x50 pixel black square. For poisoned images, the corresponding labels are modified such that source class pixels are re-labeled as the target class.

The second poisoning approach is an extension of a classification patch backdoor data poisoning attack (see for example \mycite{gu2017badnets}). We refer to this approach as a ``texture'' attack, with the goal of creating an attack that is more physically realizable than the fine-grained attack. In some fraction of training images, we replace the pixels of a designated source class with a random unseen texture image. In Fig. \ref{poison_examples}, we replace building pixels with an unseen image of foliage from US3D. Conceptually, this could represent rooftop garden spaces that many buildings are adding. While the texture attack is similar to the fine-grained attacks, it differs slightly in implementation and attacker's goal. We do not modify any of the training labels; instead we assume that the attacker is trying to take advantage of some inherent property of the dataset that causes a model to learn a potentially erroneous but exploitable representation of a class. During evaluation, the poisoned texture is then inserted randomly into images, with the goal of causing the model to classify the pixels in the inserted texture as the source class (a building). Again, in this approach, only image pixels are modified, not training labels. In practice, to ensure that the model is learning a more abstract representation of the poison texture rather than memorizing a fixed pixel pattern, we apply a series of random augmentations to the foliage image each time we replace building pixels in a poisoned training image. Specifically, the image is randomly cropped, resized, rotated, and perturbed with color jitter. 

The poisoned models were trained on 10\% poisoned data, but were otherwise trained identically to the benign (clean) models as described in \cref{training} and \cref{sec:model-dets}.

\subsection{Natural Robustness: Physically Realistic Perturbations}
\label{sec:realistic-perturb}

\begin{figure}[tb]
    \centering
     \includegraphics[width=0.8\linewidth]{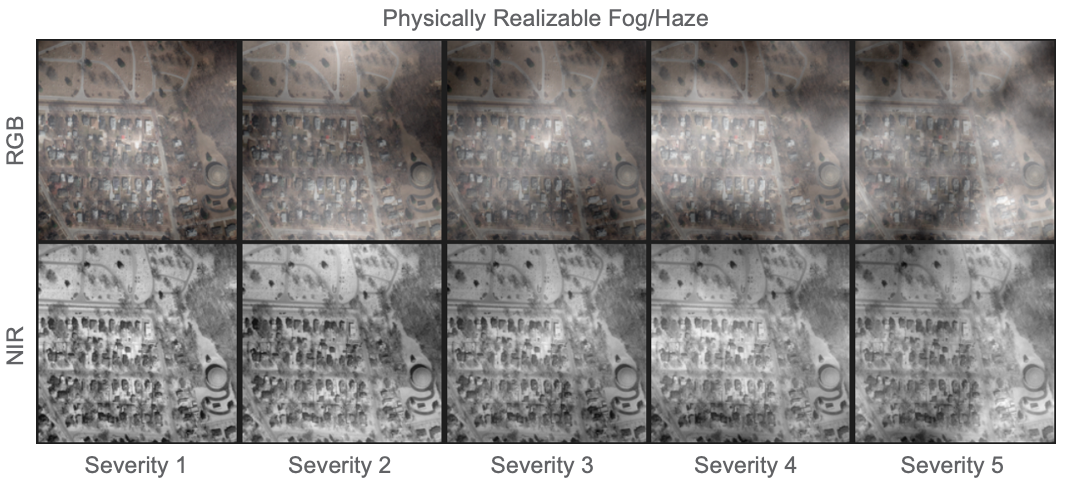}
    \caption{An example of the physically realistic fog/haze perturbations used in this work. We modify the original implementation of the ImageNet-C perturbations to account for the fact that NIR  light more easily penetrates fog, haze, and  smoke.}\label{fog_examples}
\end{figure}

For an assessment of natural robustness, we extend common corruptions presented in ImageNet-C \mycite{hendrycksBenchmarkingNeuralNetwork2019} to operate on multispectral imagery in a physically meaningful way. Specifically designed for RGB imagery, ImageNet-C is now a standard dataset used to benchmark model robustness to common corruptions. Many of the ImageNet-C corruptions model digital phenomena (e.g., shot noise, JPEG compression, contrast), which can be applied to multi-channel imagery without issue. However, the naturalistic perturbations associated with environmental conditions (e.g., snow, fog, frost) model physical processes with specific observational signatures, and additional care should be taken to ensure that such corruptions are applied to multispectral imagery in a way that faithfully captures the underlying real-world phenomena. 

There is little research on how to extend natural robustness to multispectral data. As a first step, we took two common environmental corruptions, snow and fog, and extend them to the NIR band in a physically realistic manner. We compare these modified corruptions to the original formulation to quantify any difference in performance. Finally, we use these new corruptions to assess the robustness of the different models (RGB, NIR, RGB+NIR) and fusion approaches described above. 

\textbf{Snow Corruptions}: In the implementation of snow corruptions in ImageNet-C, images are first whitened and then directional chunks of snow are added throughout. To realistically extend this to additional channels, we account for the wavelength-dependent spectral reflectance of fresh snow. In the visible R,G, and B bands, fresh snow has a reflectance of nearly 1.0, but the reflectance drops to 0.6 in the NIR band.
Accordingly, we modify the overall brightness of the snow corruptions before adding them to the NIR image, reducing the brightness of the added corruptions by 40\%. An example of our implementation can be found in \cref{sec:perturb-dets}. 

\textbf{Fog/Haze Corruptions}: 
The kind of fog applied in ImageNet-C is difficult to observe in overhead imagery. However, smoke or haze are commonly observed environmental conditions, and produce a similar visual effect as the fog in ImageNet-C in the visible bands. NIR light can more easily penetrate haze and smoke, so a realistic implementation of fog/haze should have a reduced application in the NIR band. For the fog/haze natural corruptions, we modify the overall severity of the haze added to the NIR channels similarly to the snow corruptions. Fig. \ref{fog_examples} shows an example of our implementation of fog/haze on the RGB and NIR channels. A more detailed example can be found in \cref{sec:perturb-dets}. 

\subsection{Robustness and Performance Metrics}

We use the 5 metrics defined in \mycite{liHidden2021} to quantify model performance and robustness to adversarial attacks. Mean pixel accuracies and Intersection Over Union values (IoU) are averaged across all labels, and then across all images in the unseen hold-out test split\footnote{This corresponds to \texttt{average="micro"} and \texttt{multidim\_average="global"} when using the TorchMetrics \texttt{MulticlassAccuracy} and \texttt{MultiClassJaccardIndex} classes \mycite{torchmetrics}}.
\begin{enumerate}
   \item \textbf{mIOU-B}: Benign mean IOU, calculated for a poisoned model that is evaluated on \emph{clean} data. In practice, an attacker wants to ensure that a poisoned model does not show a drop in overall performance that would be noticeable to any end users.
   \item \textbf{mPA-B}: Benign mean pixel accuracy, calculated for a poisoned model that is evaluated on \emph{clean} data. 
   \item \textbf{mIOU-A}: Attacked mean IOU, calculated for a model that is evaluated on \emph{poisoned} data.
   \item \textbf{mPA-A}: Attacked mean pixel accuracy, calculated for a model that is evaluated on \emph{poisoned} data.
   \item \textbf{ASR}: Adversarial Success Rate, calculated for a model that is evaluated on poisoned data. Measures the efficacy of an adversarial attack; higher values indicate a more successful attack. Definition varies for different poisoning attacks, as described below.
\end{enumerate}

For the fine-grained line and square attacks, ASR is the pixel-wise accuracy of the target class when the poison in present. For the example shown in Fig. \ref{poison_examples}, in an image with the square or line present, a perfect ASR score would mean that every pixel in the foliage class is predicted as the building class. In this sense, the fine-grained attack operates similarly to a targeted adversarial attack: it is not enough to simply misclassify the foliage pixels (i.e., as any other class), a successful attack must misclassify the foliage pixels as building pixels.

For the texture attack, ASR is calculated as the pixel-wise accuracy of the source class, wherever the poisoned texture appears in an image. For the example shown in Fig. \ref{poison_examples}, building pixels were replaced with the foliage texture during training, and during evaluation the foliage texture is inserted randomly into an image. A perfect ASR in this case would mean that every pixel covered by the inserted foliage texture is predicted as the building class. 

\section{Results and Evaluation}

In this section we present results for data poisoning and multispectral corruption robustness experiments.

\begin{table}[htb]
    \centering
    \begin{tabular}{||c c c c c c c c c||} 
     \hline
     Model & mIOU & mPA & ground & foliage & building & water & road & unclass\\ [0.5ex] 
     \hline\hline
     NIR & .757 & .862 & .78 & .75 & .72 & .94 & \textbf{.87} & \textbf{.45} \\
     \hline
     RGB & .761 & .865 & .78 & .75 & .74 & \textbf{.95} & \textbf{.87} & .36 \\ 
     \hline
     RGB-NIR early & \textbf{.769} & \textbf{.869}  & \textbf{.79} & \textbf{.76} & \textbf{.75} & \textbf{.95} & \textbf{.87} & .36 \\
     \hline
     RGB-NIR late & .764 & .866 & \textbf{.79} & \textbf{.76} & .74 & .92 & .86 & .36 \\ [1ex] 
     \hline
    \end{tabular}
    \caption{\textbf{Clean, unpoisoned models}: Performance metrics for clean, unpoisoned models. The class scores show IOU, and the highest score in each column is in bold. All models perform similarly across the considered metrics, with the RGB-NIR early fusion model showing the best overall performance.}\label{clean_results}
\end{table}

To establish a baseline to compare against, we first assess the performance of clean, unpoisoned models. Table \ref{clean_results} shows our clean model results, including mIOU and mPA for overall model performance and IOU scores for each class. For the inputs and fusion types considered here, all models perform similarly in all metrics. The RGB-NIR early fusion model has the best overall performance, by .003 ($0.3\%$) in pixel accuracy and .005 ($0.5\%$) in IOU. To check the significance of these results, we train five different RGB model initializations, and evaluate each on the hold-out split. We then calculated the mean and standard deviation for each metric in Table \ref{clean_results}. For pixel accuracy, we find a standard deviation of 0.001 (0.1\%). For IOU, we find a standard deviation of 0.002 (0.2\%).

\subsection{Data Poisoning}

\begin{table}[htb]
    \centering
    \begin{tabular}{|| c c c c c c ||} 
     \hline
     Model & mIOU-B & mPA-B & mIOU-A & mPA-A & Targeted ASR \\ [0.5ex] 
     \hline\hline
     NIR & .834 & .909 & .687 & .814 & \color{blue} \textbf{.921} \\
     \hline
     RGB & .841 & .913 & .691 & .817 & .924 \\ 
     \hline
     RGB-NIR early & .847 & .917 & .694 & .82 & .932 \\
     \hline
     RGB-NIR late & .843 & .915 & .691 & .817 & \color{red} \textbf{.937}\\  [1ex] 
     \hline
    \end{tabular}
    \caption{\textbf{Fine-grained line attack:} mIOU-B shows the performance of the poisoned model evaluated on benign (clean) data, while mIOU-A shows the performance of the poisoned model evaluated on attacked (poisoned) data. The red text highlights the highest ASR, while the blue text highlights the lowest ASR.}\label{line_poison}
\end{table}

\begin{table}[htb]
    \centering
    \begin{tabular}{|| c c c c c c ||} 
     \hline
     Model & mIOU-B & mPA-B & mIOU-A & mPA-A & Targeted ASR \\ [0.5ex] 
     \hline\hline
     NIR & .833 & .909 & .685 & .813 & \color{blue} \textbf{.928}\\
     \hline
      RGB & .841 & .914 & .69 & .817 & .931  \\ 
     \hline
     RGB-NIR early &  .846 & .917 & .693 & .819 &  \color{red} \textbf{.932}\\
     \hline
     RGB-NIR late & .841 & .914 & .69 & .817 & \color{blue} \textbf{.928}\\  [1ex] 
     \hline
    \end{tabular}
    \caption{\textbf{Fine-grained square attack:} Red and blue show the highest and lowest ASR scores.}\label{square_poison}
\end{table}

\begin{table}[htb]
    \centering
    \begin{tabular}{|| c c c c c c ||} 
     \hline
     Model & mIOU-B & mPA-B & mIOU-A & mPA-A & Targeted ASR \\ [0.5ex] 
     \hline\hline
     NIR & .758 & .863 &.765 & .867 & \color{blue} \textbf{.889}\\
     \hline
     RGB & .767 & .868 & .774 & .872 & .905 \\ 
     \hline
     RGB-NIR early & .77 & .87 & .773 & .872 & .901 \\
     \hline
     RGB-NIR late & .766 & .867 & .772 & .872 & \color{red} \textbf{.911} \\  [1ex] 
     \hline
    \end{tabular}
    \caption{\textbf{Physically realizable texture attack}: Red and blue show the highest and lowest ASR scores.}\label{texture_poison}
\end{table}

Our results in Tables \ref{line_poison}, \ref{square_poison} and \ref{texture_poison} demonstrate that all models are vulnerable to all types of data poisoning under consideration (both fine-grained and our physically realizable attack). No one model is significantly more robust to data poisoning than the others, and our 10 \% data poisoning attacks result in over 90 percent adversarial success rate. Interestingly, while the RGB+NIR models are the best performing models (i.e., Table \ref{clean_results}), the RGB+NIR models also have the highest ASR scores. This suggests while the extra information provided by additional bands boosts performance, it also reduces the overall adversarial robustness. In fact, based on ASR, the single-channel NIR model is the most robust to all attacks, albeit with an extremely small advantage: we find a .003 or 0.3\% improvement for fine-grained attacks and .016 or 1.6\% improvement for physically realizable attacks when compared to the respective RGB models.

It is also interesting to note that neither early or late fusion approaches show any robustness advantage. This is different than the results of \mycite{wangAdversarial2022}, where late fusion models were found to be more robust to adversarial examples, in the context of RGB+LiDAR object detection models for natural imagery. This suggests that the robustness of model fusion techniques varies with attack type (i.e., poisoning vs. digital attacks), and with model task (i.e., segmentation vs. detection).

When comparing mIOU-B and mPA-B in Tables \ref{line_poison}, \ref{square_poison}, \ref{texture_poison} (i.e., \emph{poisoned} models evaluated on clean data) to the clean mIOU and mPA scores in Table \ref{clean_results} (i.e., \emph{clean} models evaluated on clean data), more nuanced differences appear between the two different data poisoning methods. For the fine-grained attacks in Tables \ref{line_poison} and \ref{square_poison}, the poisoned models show higher benign scores relative to the clean model scores, meaning that to the victim it would appear as if the model's performance had \emph{improved} after poisoning. When the poisoned models are evaluated on poisoned data, mIOU-A and mPA-A drop significantly compared to the benign scores, by approximately 15\%. In both respects, the physically realizable texture attack shows different behavior. First, the texture attack has very similar benign and clean scores, meaning that model performance appears normal to the victim. Additionally, the attacked mIOU-A and mPA-A do not show the same sharp decrease in performance seen in the fine-grained attack (less than a percent), despite similar ASR scores. From this perspective, the physically realizable attack is a more stealthy attack than the fine-grained attack.

\subsection{Natural Perturbations}

In this section we evaluate model robustness to the physically realistic common corruptions introduced in \cref{sec:realistic-perturb}, and include evaluations of robustness to unrealistic corruptions for comparison. All models appearing below were trained with clean data. Overall, we find that input channels and fusion type \emph{do} play a role in how well models perform on our natural perturbation datasets.

\begin{figure}
    \centering
    \includegraphics[width=0.8\linewidth]{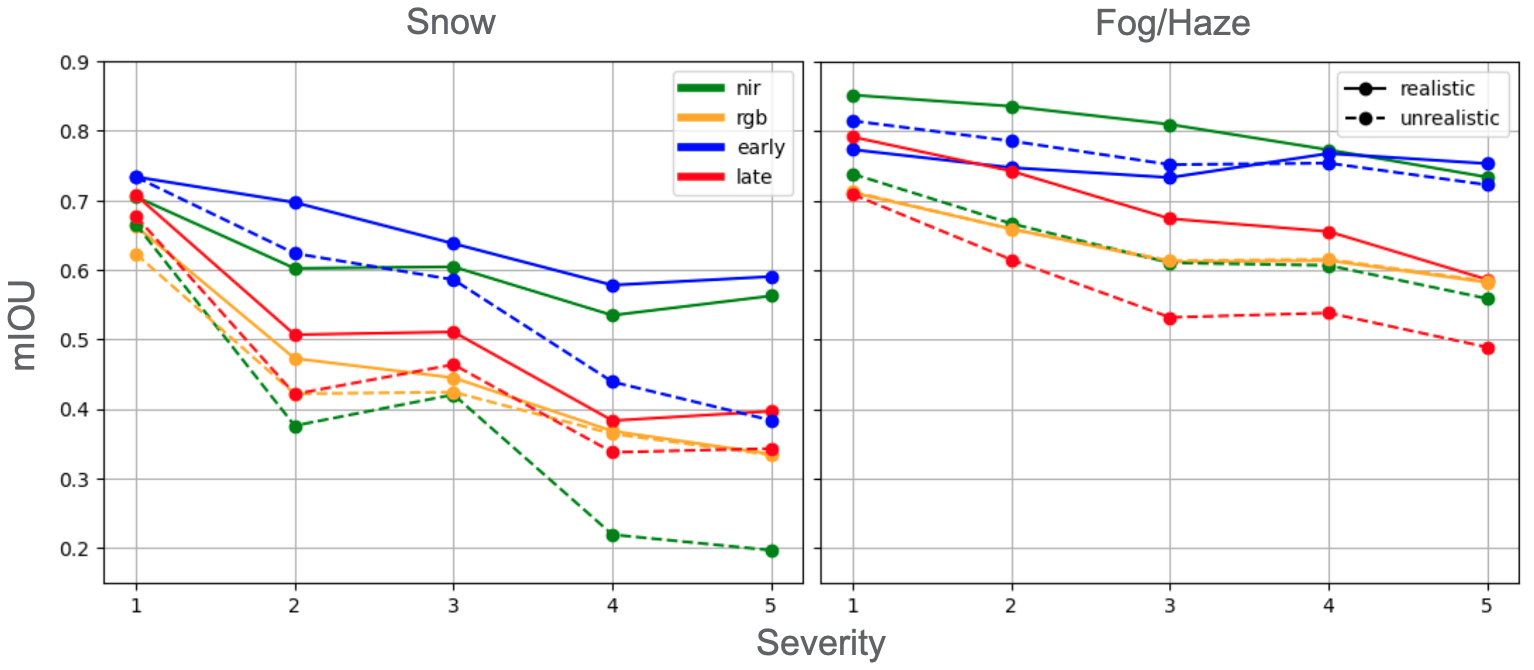}
    \caption{Model accuracy on data corrupted with physically realistic snow (left) and fog/haze (right) at varying levels of severity. In these plots solid lines denote physically realizable whereas dotted lines denote unrealistic digital (i.e. a naive extension to 4 channels). Different colors represent different model architectures. 
    % (I don't love the legend and am open to different naming and changes)
    }\label{snow_haze}
\end{figure}

Fig. \ref{snow_haze} compares mIOU scores for physically realistic perturbations (solid line) and the original, unrealistic implementation (dotted line) as a function of corruption severity. For both snow and fog/haze, models perform noticeably better on the physically realistic perturbations when compared to the original, unrealistic perturbations. This result is perhaps to be expected, given the details of our physically realistic corruption implementation, which effectively reduces the strength of the perturbations in the NIR band to better match real-world conditions (wavelength-dependent fresh snow albedo; the transparency of haze at NIR wavelengths).

On a more interesting note, for both snow and fog/haze perturbations, model performance changes with input types and fusion methods, an effect not seen in the data poisoning results. In Fig. \ref{snow_haze}, the NIR and early fusion models are more robust to the physically realistic perturbations compared to the RGB and late fusion models. This is consistent with concurrent work suggesting early fusion models rely more on NIR  than their late fusion counterparts to make decisions \mycite{us}.

\begin{table}[htb]
    \centering
    \begin{tabular}{||c c c c c c c c ||} 
     \hline
     Model & mIOU & mPA & ground & foliage & building & water & road\\ [0.5ex] 
     \hline\hline
     NIR & .60 & .75 & .64 & .63 & .42 & .29 & .26  \\
     \hline
     RGB & .46 & .62 & .55 & .27 & .30 & .52 & .24 \\ 
     \hline
     RGB-NIR early & .65 & .78  & .68 & .63 & .46 & .75 & .45  \\
     \hline
     RGB-NIR late & .50 & .66 & .60 & .44 & .27 & .18 & .19 \\ [1ex] 
     \hline
    \end{tabular}
    \caption{\textbf{Robustness to physically realistic snow perturbations:} Metrics are averaged across the 5 different corruption severities; class scores are IOU.}\label{snow_average_results}
\end{table}

\begin{table}[htb]
    \centering
    \begin{tabular}{||c c c c c c c c ||} 
     \hline
     Model & mIOU & mPA & ground & foliage & building & water & road\\ [0.5ex] 
     \hline\hline
     NIR & .67 & .80 & .70 & .61 & .56 & .82 & .76  \\
     \hline
     RGB & .47 & .64 & .56 & .30 & .22 & .38 & .43 \\ 
     \hline
     RGB-NIR early & .61 & .76  & .67 & .59 & .40 & .25 & .57  \\
     \hline
     RGB-NIR late & .53 & .69 & .62 & .51 & .21 & .22 & .17 \\ [1ex] 
     \hline
    \end{tabular}
    \caption{\textbf{Robustness to physically realistic fog/haze perturbations:} Each metric is averaged across the 5 different corruption severities; class scores are IOU.}\label{fog_average_results}
\end{table}

In Tables \ref{snow_average_results} and \ref{fog_average_results}, we focus on results from our physically realizable perturbations. These tables show \emph{overall} robustness to natural corruptions, where the metrics have now been averaged over the 5 severities shown in Fig. \ref{snow_haze}. In both Tables \ref{snow_average_results} and \ref{fog_average_results}, the NIR model shows one of the best overall natural robustness, simply confirming what we already know: that NIR imagery is more useful across varying weather and environmental conditions. As such, we ignore the NIR-only models in the remaining analysis, since these inputs saw an overall reduced level of perturbations. 

Overall, the early and late fusion models show improved robustness over the RGB model, suggesting that these models are able to leverage NIR information to improve segmentation performance in adverse weather conditions. The early fusion model shows the best overall robustness; as discussed earlier, this is in-line with research from \mycite{us} that suggests early fusion models rely more on NIR inputs.

The class-specific scores from Tables \ref{snow_average_results} and \ref{fog_average_results} reveal other interesting insights worth highlighting. Both fusion models perform better on the foliage class than the RGB model for both snow and fog/haze corruptions, with the early fusion model showing an 0.36 improvement in IOU score over the RGB model for snow (a nearly 60\% improvement) and similar improvements for fog/haze. Foliage appears quite bright at infrared wavelengths, and when paired with the results from \mycite{us} that finds that early fusion models rely more on NIR inputs, provides a reasonable explanation for the improved performance found in the early fusion models. 

\section{Conclusion}

In this paper we study the adversarial and natural robustness of multispectral segmentation models. Our main findings can be summarized accordingly:
\begin{enumerate}
    \item We find that all segmentation models are vulnerable to data poisoning attacks, regardless of input (NIR, RGB, NIR+RGB) or fusion architecture (early, late). Both the fine-grained attacks and physically realizable texture attacks are highly successful (ASR $> 90\%$) with only 10\% of the training images poisoned; however, the texture attacks are less likely to be detected by the victim.
    \item The two RGB+NIR models show the best overall performance as measured by accuracy and IOU, but \emph{also} the worst overall robustness to adversarial attacks. We conclude that the additional information provided by the additional input bands boosts overall performance, but does so at the expense of adversarial robustness.
    \item In contrast with previous work in object detection \mycite{wangAdversarial2022}, we did not find any significant difference in adversarial robustness between early and late fusion approaches, suggesting that the adversarial robustness of fusion approaches varies with attack type and model task.
    \item We create a physically realistic version of the ImageNet-C snow and fog corruptions that are appropriate for multispectral data and faithfully preserve the real-world observational signatures of snow and fog/haze.
    \item We find that both RGB+NIR models show improved robustness to natural perturbations over RGB-only models, suggesting that these models are able to successfully leverage NIR information to improve segmentation performance in adverse weather conditions. We find that the early fusion models have the best overall natural robustness, which aligns with results from \mycite{us} that find that the early fusion models rely more on NIR inputs. Additionally, the foliage class, which has a distinct NIR signature, shows significant improvement in the early fusion model.
\end{enumerate}

We leave several research directions open for future work. Extending adversarial and natural perturbations to multispectral models with additional input bands would help shed light on whether or not more channels make for less or more vulnerable models. There remain many unexplored ways to test robustness of multispectral models to physically realistic corruptions, for example isolating test splits to probe sub-optimal environmental conditions. Additionally, US3D also includes LiDAR observations matched to the multispectral images, and it would be interesting to extend the presented research to this new modality.

\acknowledgments % equivalent to \section*{ACKNOWLEDGMENTS}       
 
The research described in this paper was conducted
under the Laboratory Directed Research and Development Program at Pacific
Northwest National Laboratory, a multiprogram national laboratory operated by
Battelle for the U.S. Department of Energy.

\appendix    %>>>> this command starts appendixes

\section{IMAGE PROCESSING AND DATA SPLITS}
\label{sec:ds-dets}

US3D is a multi-modal dataset that builds upon the SpaceNet Challenge 4 dataset \mycite{SN4}(hereafter SN4). SN4 was originally designed for building footprint estimation in off-nadir imagery, and includes satellite imagery from Atlanta, GA for view angles between 7 and 50 degrees. US3D uses the subset of Atlanta, GA imagery from SN4 for which there is matched LiDAR observations, and adds additional matched satellite imagery and LiDAR data in Jacksonville, FL and Omaha, NE. The Atlanta imagery is from Worldview-2, with ground sample distances (GSD) between 0.5m and 0.7m, and view angles between 7 and 40 degrees. The Jacksonville and Omaha imagery from Worldview-3, with GSD between 0.3m and 0.4m, and view angles between 5 and 30 degrees. As described below, we train and evaluate models using imagery from all three locations. We note however, that models trained solely on imagery from a single location will show variation in overall performance due to the variations in the scenery between locations (e.g., building density, seasonal changes in foliage and ground cover). 

The US3D dataset includes both 8-bit RGB satellite imagery and 16-bit pansharpened 8-band multispectral imagery. One of the goals of this work is to assess the utility of including additional channels as input to image segmentation models (e.g., near-infrared channels). In order to include channels beyond Red, Green, or Blue, we must work from the 16-bit pansharpened 8-band images. We briefly describe our process for creating 8-bit, 8-band imagery, which consists of rescaling, contrast stretching, and gamma correcting the pixels in each channel independently. Specifically, the original 16-bit pixel values are rescaled to 8-bit, and a gamma correction is applied using $\gamma=2.2$. The bottom 1\% of the pixel cumulative distribution function is clipped, and the pixels are rescaled such that the minimum and maximum pixel values in each channel are 0 and 255. 

We note that when applied to the R, G, and B channels of the multispectral image products to generate 8-bit RGB images, this process produces images that are visually similar but \emph{not} identical to the RGB images provided in US3D. As such, the RGB model presented in this work cannot be perfectly compared to models published elsewhere trained on the RGB imagery included in US3D\footnote{We trained identical models on the US3D RGB images and the RGB images produce in this work and found that the US3D models performed slightly better, 1-2\% improvement in average pixel accuracy. This is likely due to more complex and robust techniques used for contrast stretching and edge enhancement in US3D; unfortunately these processing pipelines are often proprietary and we could not find any published details of the process.}. However, we felt that this approach provided the most fair comparison of model performance for different input channels, since the same processing was applied identically to each channel.

Satellite images are generally quite large (hundreds of thousands of pixels on a side) and must be broken up into smaller images in order to be processed by a deep learning model, a process sometimes called ``tiling.'' Each of the large satellite images (and matched labels) were divided into 1024 pixel x 1024 pixel tiles without any overlap, producing 27,021 total images or ``tiles''. All tiles from the same parent satellite image are kept together during the generation of training and validation splits to avoid cross contamination that could artificially inflate accuracies\footnote{We note this is different from the data split divisions within US3D, which mixes tiles from the same parent image between training, validation, and testing.}. An iterative approach was used to divide the satellite images into training, validation, and hold-out (test) data splits to ensure that each data split includes imagery with consistent metadata properties: location (Atlanta, Jacksonville, Omaha), view-angle, and azimuth angle. The final data splits included 21,776 tiles in training (20\%), 2,102 tiles in validation (8\%), and 3,142 tiles in the unseen, hold-out test split (12\%).

Models with near-infrared (NIR) input use the WorldView NIR2 channel, which covers 860-1040nm. The NIR2 band is sensitive to vegetation but is less affected by atmospheric absorption when compared with the NIR1 band.

\section{MULTISPECTRAL SEGMENTATION MODEL TRAINING}
\label{sec:model-dets}

All models are trained using PyTorch \mycite{pytorch} using distributed data parallel with effective batch size of 32 (8 GPUs \(\times\) 4 datapoints per GPU), using a Dice loss function \mycite{dice} and the Adam optimizer \mycite{adam}. We use a ``reduce on plateau'' learning rate scheduler, with an initial learning rate of $10^{-3}$, a minimum learning rate of $10^{-6}$ and a Learning away drop by a factor of \(10\) when 10 epochs elapse without a \(1\%\) in validation intersection-over-union (IoU). With this learning rate scheduler, the models typically trained for 150-180 epochs before reaching the minimum learning rate.

% data augmentation? GPU used? hyperparemter tuning?

\section{PHYSICALLY REALIZABLE PERTURBATION EXAMPLES}
\label{sec:perturb-dets}

\begin{figure}[tb]
    \centering
     \includegraphics[width=0.8\linewidth]{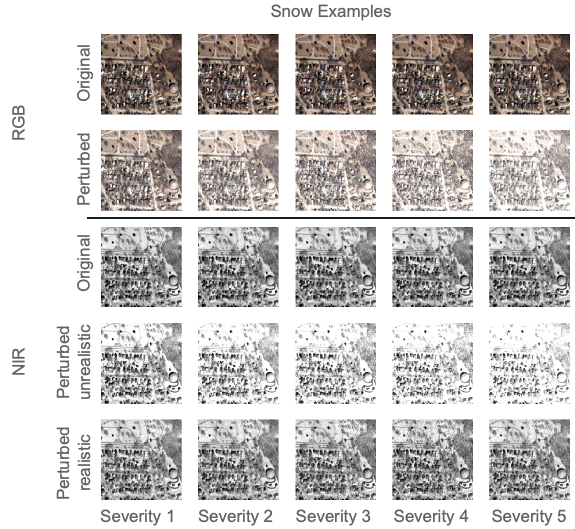}
    \caption{Comparison of our implementation (perturbed realistic) vs. \mycite{hendrycksBenchmarkingNeuralNetwork2019} (perturbed unrealistic) perturbations for snow. }\label{snow_examples}
\end{figure}

\begin{figure}[tb]
    \centering
     \includegraphics[width=0.8\linewidth]{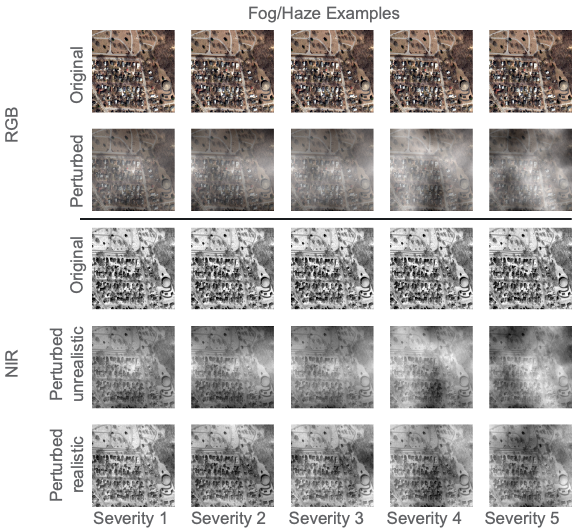}
    \caption{Comparison of our implementation (perturbed realistic) vs. \mycite{hendrycksBenchmarkingNeuralNetwork2019} (perturbed unrealistic) perturbations for fog/haze.}\label{fog_haze_examples}
\end{figure}

As discussed above, we extend \mycite{hendrycksBenchmarkingNeuralNetwork2019} snow and fog common corruptions from ImageNet-C to multispectral data. We modify the code available at \href{https://github.com/hendrycks/robustness}{github.com/hendrycks/robustness}. Fig. \ref{snow_examples} and \ref{fog_haze_examples} show examples for snow and fog on US3D. For these figures, we show unrealistic and realistic perturbations for the NIR channel. The RGB channel corruptions are already realistic, and thus remain identical to the original ImageNet-C implementation. ``Perturbed unrealistic`` refers to applying the ImageNet-C RGB corruptions to the NIR band without taking physical constraints into consideration. ``Perturbed realistic`` refers to our modifications that bring the NIR corruptions in line with real-world observations. 

\clearpage
% References
\bibliography{report} % bibliography data in report.bib
\bibliographystyle{spiebib} % makes bibtex use spiebib.bst

\end{document}